\documentclass[conference]{IEEEtran}
\IEEEoverridecommandlockouts
\usepackage{amsmath,amssymb,amsfonts}
\usepackage{algorithmic}
\usepackage{graphicx}
\usepackage{textcomp}
\usepackage{xcolor}
\usepackage[numbers]{natbib}
\usepackage{array}
\usepackage{url}
\usepackage{hyperref}
\usepackage{cleveref}
\def\BibTeX{{\rm B\kern-.05em{\sc i\kern-.025em b}\kern-.08em
    T\kern-.1667em\lower.7ex\hbox{E}\kern-.125emX}}

\begin{document}

\title{COMPARE: A 
Taxonomy and Dataset of Comparison Discussions in Peer Reviews}

\author{\IEEEauthorblockN{Shruti Singh}
\IEEEauthorblockA{\textit{Dept. of Computer Science and Engg.} \\
\textit{Indian Institute of Technology,}\\ \textit{Gandhinagar}\\
Gujarat, India \\
singh\_shruti@iitgn.ac.in}
\and
\IEEEauthorblockN{Mayank Singh}
\IEEEauthorblockA{\textit{Dept. of Computer Science and Engg.} \\
\textit{Indian Institute of Technology,}\\ 
\textit{Gandhinagar}\\
Gujarat, India \\
singh.mayank@iitgn.ac.in}
\and
\IEEEauthorblockN{Pawan Goyal}
\IEEEauthorblockA{\textit{Dept. of Computer Science and Engg.} \\
\textit{Indian Institute of Technology,}\\
\textit{Kharagpur}\\
West Bengal, India \\
pawang.iitk@gmail.com}
}

\maketitle

\begin{abstract}
Comparing research papers is a conventional method to demonstrate progress in experimental research. We present COMPARE, a taxonomy and a dataset of comparison discussions in peer reviews of research papers in the domain of experimental deep learning. From a thorough observation of a large set of review sentences, we build a taxonomy of categories in comparison discussions and present a detailed annotation scheme to analyze this. Overall, we annotate 117 reviews covering 1,800 sentences. We experiment with various methods to identify comparison sentences in peer reviews and report a maximum F1 Score of 0.49. 
We also pretrain two language models specifically on ML, NLP, and CV paper abstracts and reviews to learn informative representations of peer reviews. The annotated dataset and the pretrained models are available at \href{https://github.com/shruti-singh/COMPARE}{https://github.com/shruti-singh/COMPARE}.
\end{abstract}


\section{Introduction}
\label{sec:intro}
The advent of open-access, online peer reviewing platform OpenReview\footnote{\label{OR}https://openreview.net/} platform provides an opportunity to curate and analyse peer reviews and enhance our understanding of paper acceptance decisions. A crucial factor impacting a research paper's acceptance to a conference or a journal is its positioning with respect to the existing literature in the field. The discussion of previous research includes many aspects such as demonstrating potential challenges in previous works, comparison against state-of-the-art papers, discussion of datasets, and the evolution of techniques over time. While improvement in benchmarks/results is not the only measure of progress, it is a significant indicator of progression in a positive direction. All comparisons must be complete, precise, and thorough (hereafter, \textit{`meaningful comparison'}). In contrast, lack of a proper comparison with important previous works can be termed a \textit{`non-meaningful comparison'}.  Although meaningful comparison of papers is not the only criterion for paper acceptance to a venue, it is an indicator of the paper's quality and a primary reason for most rejected papers~\cite{Chakraborty2020AspectbasedSA}. 

\begin{figure}[!t]
    \centering
    \begin{tabular}{c}
         \underline{Candidate Paper I} \\
         \includegraphics[width=0.7\columnwidth]{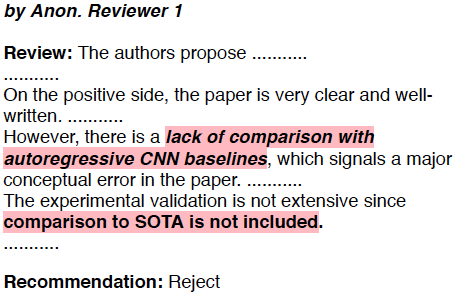}\\
          \underline{Candidate Paper II} \\
         \includegraphics[width=0.7\columnwidth]{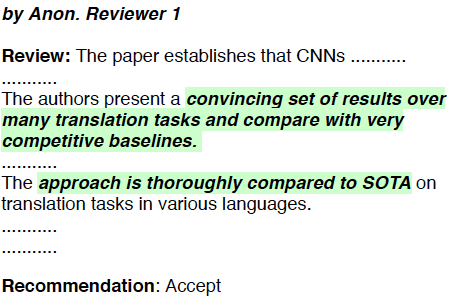}\\
    \end{tabular}
    
    \caption{Representative peer reviews containing comparative discussions. As highlighted, candidate paper I lack meaningful comparison, whereas candidate paper II contains thorough meaningful comparative experiments.}
    \label{fig:mcompexample}
\end{figure}

As per the current practice, reviewers use their domain knowledge to evaluate whether a candidate paper meaningfully compares itself against existing literature. An automated system to detect if the paper meaningfully compares against existing literature can assist the reviewers and speed-up the overall peer-review pipeline. However, several challenges exist in developing such predictive systems: (i) the volume and diversity of existing literature and (ii)  meaningful comparison is a highly subjective and multi-faceted problem. For example, a candidate paper can be rejected if it only compares itself against weaker baselines or presents comparative results on non-standard,  small-scale, or biased datasets. Peer review texts often evaluate and discuss the comparative analysis conducted in the candidate paper. Figure~\ref{fig:mcompexample} illustrates examples of review texts discussing the meaningful and non-meaningful comparison of papers. As a first step towards understanding the various aspects of comparison discussions, we propose an exhaustive hierarchical categorization of different scenarios from comparison discussions in peer review texts (\Cref{sec:hierarchy}). 

In this paper, we present \textit{COMPARE}, a taxonomy and a manually curated dataset of comparative discussions present in peer reviews of papers submitted to \textit{International Conference on Learning Representations (ICLR)} between the years 2017--2020. We leverage the peer reviews to build a taxonomy of comparison discussion consisting of four aspects and thirteen subcategories (Section~\ref{sec:hierarchy}). The dataset (Section~\ref{sec:dataset}) comprises $\sim$1,800 sentences with comparison and non-comparison labels. We also annotate each sentence with fine-grained categories as per our proposed hierarchical categorization. We pretrain two language models based on RoBERTa~\cite{Liu2019RoBERTaAR}: (i) MLRoBERTa, and (ii) MLEnRoBERTa, specifically on ML, NLP, and CV paper abstracts, and ICLR review texts (Section \ref{sec:encoders}). 
MLEnRoBERTa is trained on the masked dataset, in which scientific entities are masked with \textit{Task, Material, Method}, and \textit{Metric} labels.
We experiment with several ML models (Section~\ref{sec:methods}) to categorize a peer-review sentence into either comparative or non-comparative class.

\section{Related Work}
\label{sec:rel_work}
Several works focus on curation of peer review data~\cite{Kang2018ADO}, and further analysis such as sentiment analysis of reviews and decision prediction~\cite{Kang2018ADO, Ghosal2019DeepSentiPeerHS}.
Several works~\cite{hua-etal-2019-argument-mining,Fromm2020ArgumentMD} focus on extraction and prediction of diverse argumentation schemes such as: (i) Evaluation, Request, Fact, Reference, Quote, and Non-argument, and (ii) Non-argument, Supporting arguments, and Attacking arguments.
\citet{Fortanet2008EvaluativeLI} analyse peer reviews in Applied Linguistics and Business Organisation fields and propose a broad taxonomy of reviews and present linguistic patterns for three categories: Crticism, Recommendation, and Requests. \citet{gosden2003not} provide a taxonomy consisting of categories: Technical Detail, Claims, Discussion, References, and Format. \citet{mungra2010peer} identify frequent linguistic patterns in works submitted to medico-scientific journals by Italian researchers by using categories and data-terms from previous works~\cite{gosden2003not,day2002use}.

\citet{Chakraborty2020AspectbasedSA} annotate $\sim$2,500 review sentences with eight aspects, including the aspect meaningful comparison, and their associated sentiments.
Since~\citet{Chakraborty2020AspectbasedSA} choose sentences closest to the seed set of sentences in terms of cosine similarity as candidates for annotation, the dataset does not exhaustively cover all scenarios of comparison discussions.
To the best of our knowledge, no prior work presents an elaborate discussion on `meaningful comparison' of ML research papers and the associated hierarchical structure. Also, none of the previous works curate a comparison-specific dataset of research papers and propose approaches to automatically identify comparison sentences from the data.

\section{The Taxonomy of Comparisons}
\label{sec:hierarchy}
\begin{figure}[!tbh]
\centering
\includegraphics[width=0.88\columnwidth]{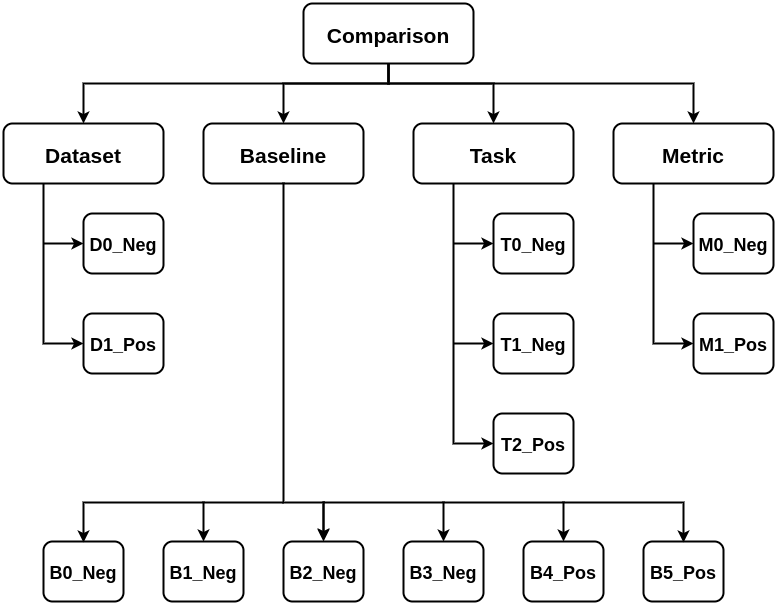}
\caption{Taxonomy of comparative review sentences.}
\label{fig:ann_classes}
\end{figure}

A thorough observation of a large volume of review sentences suggests four comparison categories: (i) dataset, (ii) baseline, (iii) task, and (iv) metric-specific.
Each of the four categories is further divided into subcategories. Each subcategory is associated with a positive or negative sentiment, and hence each comparison sentence also contains a sentiment label. Positive and negative sentiment subcategory denote meaningful-comparison and non-meaningful comparison respectively. We denote the sentiment of the subcategory by the suffix of the subcategory name. 
Figure~\ref{fig:ann_classes} details the proposed taxonomy. Due to space limitations, exemplar sentences for each subcategory are presented on \href{https://github.com/shruti-singh/COMPARE}{https://github.com/shruti-singh/COMPARE}.

\begin{enumerate}
    \item \textbf{Dataset}: Dataset-specific category sentences outline comparative discussions around the dataset used in the candidate paper. The two subcategories are:
    \begin{itemize}
        \item \emph{\textbf{D0\_Neg}}: Candidate paper requires a thorough evaluation over one or more standard datasets. 
        \item \emph{\textbf{D1\_Pos}}: Candidate paper has evaluated itself on a sufficiently relevant number of datasets. 
    \end{itemize}
    \item \textbf{Baseline}: Baseline-specific sentences discuss the completeness, thoroughness, and relevancy of baseline papers cited in the candidate paper. The baseline-specific category is divided into six sub-categories:
    \begin{itemize}
        \item \emph{\textbf{B0\_Neg}}: Candidate paper requires comparison against more state-of-the-art baselines. 
        \item \emph{\textbf{B1\_Neg}}: Candidate paper requires comparison against an explicitly stated baseline. In contrast to B1\_Neg, B0\_Neg category sentences are general recommendations.
        \item \emph{\textbf{B2\_Neg}}: Candidate paper inadequately compares itself against a specific baseline.
        \item \emph{\textbf{B3\_Neg}}: Candidate paper compares itself against incomparable baselines (a.k.a \textit{unfair comparison}). 
        \item \emph{\textbf{B4\_Pos}}: Candidate paper has compared against a specific baseline. 
        \item \emph{\textbf{B5\_Pos}}: Candidate paper has compared against all possible baselines. 
    \end{itemize}

    \item \textbf{Task}: Task-specific sentences discuss the merits of the proposed task in the candidate paper. This category is divided into three subcategories:
    \begin{itemize}
        \item \emph{\textbf{T0\_Neg}}: Candidate paper requires a comparison of results on one or more tasks.
\item \emph{\textbf{T1\_Neg}}: Candidate paper requires a comparison on a specific task. 
        \item \emph{\textbf{T2\_Pos}}: Reviewers affirm that the paper compares exhaustively on sufficient tasks. 

    \end{itemize}
    \item \textbf{Metric}:  Metric-specific sentences measure the relevancy of evaluation metrics employed in the candidate paper. This category is divided into two subcategories:

    \begin{itemize}
        \item \emph{\textbf{M0\_Neg}}: The candidate paper uses an unfair comparison metric, or a specific metric is missing. 
        \item \emph{\textbf{M1\_Pos}}: The comparison metrics used by the candidate paper are fair or exhaustive.
    \end{itemize}
\end{enumerate}

\section{Dataset}
\label{sec:dataset}
We use the peer review texts publicly available on the OpenReview\footnotemark[\getrefnumber{OR}]. We select 39 papers (with equal proportions of accepted and rejected decisions, evenly spread out across four years) submitted to ICLR between 2017--2020. 
Overall, we sample 117 review texts\footnote{Each paper receives atleast two and atmost four reviews. Most of the papers ($>$90\%) contain three review texts.}. For each review text, we manually\footnote{The first author performs the annotation.} 
classify each sentence as comparison or non-comparison. Note that the annotation task is not trivial as it requires domain-specific expertise. Overall, we annotate 143 comparison sentences and 1,658 non-comparison sentences. Each comparison sentence is further annotated into four broad categories and associated subcategories (described in Section~\ref{sec:hierarchy}). 

\begin{table}[!t]
\caption{Statistic of manual annotation task. }
\centering
\small{
\begin{tabular}{|c|c|c|}
\hline
 & \textbf{Comparison} & \textbf{Non-comparison}\\ \hline
\textbf{ Accepted} & 48 & 644 \\ \hline
\textbf{ Rejected} & 69 & 744 \\ \hline
\textbf{Total} & 143 & 1,658 \\ \hline
\end{tabular}}
\label{table:comp}
\end{table}

\begin{table}[h]
\caption{Aspect distribution of comparison sentences. Columns A0-A5 show the distribution of comparison sentences into subcategories. E.g.: \textit{A0} for \textit{Task} denotes subcategory T0\_Neg.}
\centering
\newcolumntype{A}{>{\centering\arraybackslash}m{0.08\columnwidth}}
\newcolumntype{C}{>{\centering\arraybackslash}m{0.16\columnwidth}}
\newcolumntype{P}{>{\centering\arraybackslash}m{0.03\columnwidth}}
\small{
\begin{tabular}{|C|A|P|P|P|P|P|P|}
\hline
\textbf{Aspect} & \textbf{Count} & A0 & A1 & A2 & A3 & A4 & A5 \\ \hline
Dataset  & 32 & 26 & 6 & - & - & - & - \\ \hline
Baseline  & 101 & 53 & 27 & 8 & 9 & 2 & 3 \\ \hline
Task  & 13 & 7 & 1 & 5 & - & - & - \\ \hline
Metric  & 3 & 2 & 1 & - & - & - & - \\ \hline
\end{tabular}
}
\label{table:aspcomp}
\end{table}

Table~\ref{table:comp} provides statistics of the annotated dataset. Baseline-specific comparisons are the most frequent category in the dataset, with 101 comparison sentences. Dataset-specific category contains 32 sentences. The other two categories, task-specific and metric-specific, yield 13 and 3 sentences, respectively. We argue that most of the papers use fair evaluation metrics and hence, receive fewer metric-specific suggestions. A small fraction of sentences ($\sim$5\%) can be assigned to multiple broad categories. In such cases, we assign multiple categories to the sentence. Table~\ref{table:aspcomp} presents the statistics of the number of comparison sentences in each aspect category and subcategory. 

\section{Classification Approaches}
\label{sec:methods}
We experiment with various classifiers 
to identify comparison sentences in our dataset. We randomly split the annotated dataset into the train and test sets. The train set comprises 21 reviews\footnote{from four rejected papers and three accepted paper}. The train set comprises 296 review sentences, out of which 26 sentences are labeled as comparative sentences. The remaining annotated dataset comprising 96 reviews is considered as a \textit{test} set. Next, we describe several encoding schemes to encode a review sentence before classification.

\begin{table}[!t]
\caption{Precision (P), Recall (R) and F1 scores for comparative sentences classification. SR refers to sentence representations. CR refers to chunk representations.}
\centering
\small{
\resizebox{\hsize}{!}{
\begin{tabular}{|c|c|c|c|c|}
\hline
& \textbf{P} & \textbf{R} & \textbf{F1} & \textbf{Best} \\ \hline
USE + SR & 0.42 & 0.50 & 0.45 & GNB \\ \hline
BERT + SR & 0.41 & 0.21 & 0.27 & LR \\ \hline
SciBERT + SR & \textbf{0.59} & 0.32  & 0.42 & LR  \\ \hline
MLRoBERTa + SR & 0.37 & 0.41 & 0.39 & GNB \\ \hline
MLEnRoBERTa + SR & 0.36 & 0.45 & 0.40 & GNB \\ \hline
USE + CR & 0.50 & 0.44 & 0.47 & GNB \\ \hline
BERT + CR & 0.42 & 0.33 & 0.37 & LR \\ \hline
SciBERT + CR & 0.51 & 0.34 & 0.41 & LR \\ \hline
MLRoBERTa + CR & 0.43 & 0.58 & \textbf{0.49} & GNB \\ \hline
MLEnRoBERTa + CR & 0.40 & \textbf{0.61} & 0.48 & GNB \\ \hline
\end{tabular}}}
\label{tab:classifier}
\end{table}

\subsection{Encoding peer review sentences}
\label{sec:encoders}
A key component in our classification pipeline is the language model used to encode the peer review text. We experiment with three pre-trained models: Universal Sentence Encoder (USE,~\cite{cer2018universal}), BERT (base uncased,~\cite{devlin-etal-2019-bert}), SciBERT~\cite{Beltagy2019SciBERTAP}. We also experiment with two variants of the RoBERTa~\cite{Liu2019RoBERTaAR} model. Both variants are trained on the abstracts of accepted papers from 
NeurIPS\footnote{https://papers.nips.cc/} (1987--2019), CVPR\footnote{ttps://openaccess.thecvf.com/menu} (2013--2020), ICLR\footnote{\label{ICLRFT}https://openreview.net/group?id=ICLR.cc} (2016--2020), the entire ACL Anthology dataset \cite{radev2013acl}, and ICLR paper reviews\footnotemark[\getrefnumber{ICLRFT}] (2017--2020).
The first variant (hereafter, \textit{`MLEnRoBERTa'}) uses a masked version of the above dataset. We use a curated list of scientific 
entities~\cite{houyufang2019acl,jain-etal-2020-scirex}
to mask n-grams with labels as \textit{Task}, \textit{Material}, \textit{Method}, and \textit{Metric}. The second RoBERTa variant (hereafter, \textit{`MLRoBERTa'}) does not use the masked dataset. 

Given a review sentence and the above encoding models, we represent the sentence in two possible ways. In the first case, we encode the entire peer review sentence directly without any preprocessing. In the second case, we encode chunks that contain keywords like \textit{method, task, dataset, metric}, and \textit{baseline}. If no such chunk can be extracted, we encode the whole sentence. The exact formulation of the chunking approach is skipped due to space constraints.

\subsection{Classification using ML classifiers}
We encode sentences in the training set using text encoders (Section~\ref{sec:encoders}) to train 10 traditional ML classifiers: (i) SVM with linear, polynomial, and RBF kernel (SVM), (ii) Logistic Regression with L1 and L2 regularization (LR), (iii) Decision Tree (DT), (iv) Random Forest (RF), (v) Gaussian Naive Bayes (GNB), (vi) Gaussian Process Classifier (GPC), (vii) K-Nearest Neighbors (KNN), (viii) AdaBoost (AB), (ix) Bagging Classifier (BC), and (x) Multilayer Perceptron (MLP). We use 5-fold cross-validation and grid search for hyperparameter tuning.

\section{Classification Results}
Table~\ref{tab:classifier} details the classification results on the test set. We observe maximum precision with the SciBERT model with sentence representations. However, SciBERT yields a lower recall value than other encoding schemes. We observe a better F1 score with chunk representations than sentence representations with all models except for SciBERT, which performs similarly with both the encoding schemes. Both MLRoBERTa and MLEnRoBERTa, perform better than SciBERT in terms of F1 score when paired with chunk representations. One thing to be mindful of while drawing conclusions from the results of the ML classifiers is that the dataset is imbalanced and small with only 8\% comparative sentences. We are currently extending the proposed dataset by leveraging semi-supervised labeling techniques.

\section{Conclusion and Future Work}
\label{sec:conc}
To the best of our knowledge, this is the first work to study meaningful comparison discussions.
We believe that automatic extraction and analysis of comparison discussions in peer review texts will extensively enhance our understanding of factors that affect the paper acceptance decisions. The automatic summarization of comparative statements from different reviewers can help meta-reviewers quickly glance over the concise discussion on the paper's overall experimental results comparison. Furthermore, this dataset can be used for the problem where given the candidate paper text, the system generates comparative statements mentioning if the paper has performed meaningful comparison. The hierarchy of comparison can also be used for evaluation of papers by reviewers.

\bibliographystyle{IEEEtranN}
\bibliography{IEEEabrv,jcdl2021}

\end{document}